\documentclass[conference]{IEEEtran}
\IEEEoverridecommandlockouts

\usepackage{cite}
\usepackage{amsmath,amssymb,amsfonts}
\usepackage{algorithmic}
\usepackage{graphicx}
\usepackage{textcomp}
\usepackage{xcolor}
\usepackage{booktabs}
\usepackage{array}
\usepackage{url}
\usepackage{balance}
\usepackage{tikz}
\usetikzlibrary{arrows.meta,positioning,calc,matrix,fit}
\usepackage{pgfplots}
\pgfplotsset{compat=1.18}
\def\BibTeX{{\rm B\kern-.05em{\sc i\kern-.025em b}\kern-.08em
    T\kern-.1667em\lower.7ex\hbox{E}\kern-.125emX}}

\begin{document}

\title{Experimental Analysis of Neural Network-Based Image Classification on the CIFAR-10 Dataset}

\author{
\IEEEauthorblockN{Necati Kagan Erkek, Emre Balci and Berkin Halay}
\IEEEauthorblockA{Department of Electronics and Communication Engineering, Istanbul Technical University, Istanbul, Turkey}
\IEEEauthorblockA{Email: \texttt{erkek17@itu.edu.tr, balcie17@itu.edu.tr, halay18@itu.edu.tr,}}
}

\maketitle

\begin{abstract}
An experimental investigation of neural image classification on the CIFAR-10 benchmark is presented through fully connected and convolutional network formulations. The analysis emphasizes the complete learning pipeline: image vectorization, normalization, one-hot class encoding, supervised loss minimization, learning-rate selection, mini-batch training, convolutional feature extraction, max-pooling, and validation-based generalization assessment. A convolutional architecture with six convolutional layers and three max-pooling stages is evaluated for ten training epochs using a batch size of 128 and an Adam optimizer with a learning rate of 0.001. The validation accuracy reaches approximately 74.77\%, while the validation loss begins to increase after the middle of training despite continued reduction in training loss. The resulting behavior illustrates the practical difference between representation learning and memorization, and it provides a compact experimental baseline for future studies on regularization, data augmentation, deeper architectures, and reproducible image-classification education.
\end{abstract}

\begin{IEEEkeywords}
CIFAR-10, convolutional neural network, multilayer perceptron, image classification, supervised learning, overfitting, hyperparameter tuning.
\end{IEEEkeywords}

\section{Introduction}
Image classification remains a central problem in computer vision because a classifier must transform raw pixels into semantic object labels under illumination, pose, background, and intra-class variation. Artificial neural networks offer a data-driven solution by learning nonlinear mappings between image measurements and target categories. Early multilayer perceptrons (MLPs) demonstrated the ability of back-propagation to optimize many connected parameters \cite{rumelhart1986learning}. However, dense networks do not contain an explicit spatial prior and therefore treat neighboring pixels in the same manner as distant pixels after vectorization. Convolutional neural networks (CNNs) address the limitation through local receptive fields, weight sharing, and pooling operations, making them particularly suitable for images \cite{lecun1998gradient,lecun2015deep}.

The CIFAR-10 benchmark provides a compact yet challenging environment for evaluating neural image classifiers \cite{krizhevsky2009learning}. Each image has only $32\times32$ pixels, causing fine details to be difficult to identify even by visual inspection. The low resolution produces an appropriate experimental setting for observing how representation learning, optimization, and validation behavior interact. A dense MLP can be used to expose the consequences of flattening the image into a vector, whereas a CNN can preserve local spatial information during feature extraction. The comparison is therefore pedagogically valuable as well as experimentally relevant.

The presented investigation is designed as an academic reconstruction of an image-classification experiment using the IEEE conference format. The scope includes a formal description of the dataset, the theoretical basis of MLP and CNN classifiers, the learning objective, optimizer behavior, hyperparameter effects, and result interpretation. Special attention is given to the relation between training loss and validation loss because that relation determines whether additional epochs are likely to improve generalization or merely increase memorization. The experimental log indicates that validation accuracy stabilizes near 75\%, while validation loss increases after an initial decline. Such behavior is a practical example of overfitting and motivates future studies involving dropout, batch normalization, augmentation, and deeper residual architectures \cite{srivastava2014dropout,ioffe2015batch,he2016deep}.

The broader importance of the experiment lies in its reproducibility and extensibility. A small benchmark can be executed with limited computational resources, yet it exposes the same methodological questions encountered in larger visual-recognition systems: data partitioning, objective selection, learning-rate control, capacity selection, validation monitoring, and generalization analysis. Consequently, the experiment can serve as a baseline for future studies on robust low-resolution recognition, architecture search, educational laboratory design, and comparative evaluations of training strategies.

\section{Dataset and Pre-processing}

\subsection{CIFAR-10 Image Collection}

\begin{figure}[t]
\centering
\includegraphics[width=0.86\columnwidth]{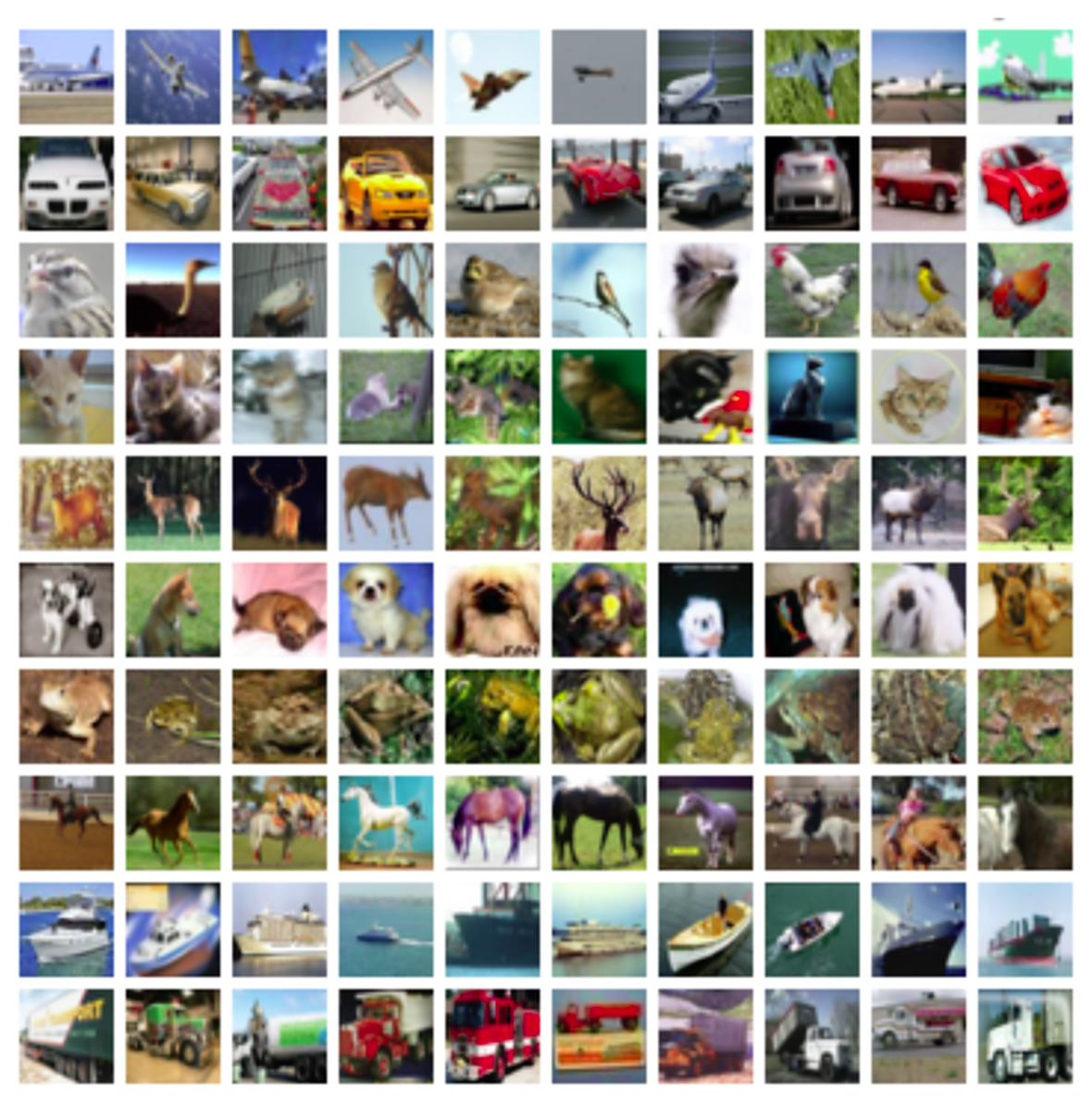}
\caption{Representative CIFAR-10 samples after visual enhancement of the supplied image grid. The small spatial resolution creates ambiguity, making the benchmark useful for comparing dense and convolutional representations.}
\label{fig:cifar_grid}
\end{figure}

CIFAR-10 contains 60,000 color images distributed over ten balanced classes: airplane, automobile, bird, cat, deer, dog, frog, horse, ship, and truck \cite{krizhevsky2009learning}. The official partition includes 50,000 training images and 10,000 test images. In the experiment, 5,000 samples from the training partition are reserved for validation, while the remaining samples are used for parameter learning. Balanced labels are important because accuracy is then interpretable without major correction for class imbalance. A sample grid is shown in Fig.~\ref{fig:cifar_grid}. The figure also illustrates the difficulty of assigning object categories from small images, especially for animal classes and visually similar vehicle classes. Each RGB image is represented by three channels with integer pixel intensities in $[0,255]$. For an image $I_i\in\{0,\ldots,255\}^{32\times32\times3}$, preprocessing maps the data into a normalized tensor:
\begin{equation}
X_i = \frac{I_i}{255}, \qquad X_i\in[0,1]^{32\times32\times3}
\label{eq:normalization}
\end{equation}

Normalization improves numerical conditioning by keeping input magnitudes within a range compatible with gradient-based optimization. For the MLP baseline, the tensor is reshaped into a vector $x_i\in\mathbb{R}^{3072}$, where $3072=32\cdot32\cdot3$. For the CNN model, the spatial tensor form is preserved so that convolutional filters can exploit two-dimensional locality.

\subsection{Label Encoding and Data Splits}
The target label for each sample is encoded as a one-hot vector $y_i\in\{0,1\}^{10}$ with exactly one nonzero element. If class $c$ is the correct category, the target vector satisfies
\begin{equation}
 y_{i,k}=\begin{cases}
 1, & k=c,\\
 0, & k\ne c,
 \end{cases}
 \qquad \sum_{k=1}^{10} y_{i,k}=1.
\label{eq:onehot}
\end{equation}
The output layer also contains ten units, and each unit corresponds to a class. Table~\ref{tab:encoding} summarizes the representation used by the supervised classifier. The validation subset is not used to update network parameters. Instead, it supplies an unbiased intermediate estimate for hyperparameter and epoch selection. The test subset is reserved for final generalization assessment after model selection.

\begin{table}[t]
\caption{CIFAR-10 Class Encoding Used by the Output Layer}
\label{tab:encoding}
\centering
\footnotesize
\begin{tabular}{@{}clc@{}}
\toprule
Index & Class & Target pattern \\
\midrule
0 & Airplane & $[1,0,0,0,0,0,0,0,0,0]$ \\
1 & Automobile & $[0,1,0,0,0,0,0,0,0,0]$ \\
2 & Bird & $[0,0,1,0,0,0,0,0,0,0]$ \\
3 & Cat & $[0,0,0,1,0,0,0,0,0,0]$ \\
4 & Deer & $[0,0,0,0,1,0,0,0,0,0]$ \\
5 & Dog & $[0,0,0,0,0,1,0,0,0,0]$ \\
6 & Frog & $[0,0,0,0,0,0,1,0,0,0]$ \\
7 & Horse & $[0,0,0,0,0,0,0,1,0,0]$ \\
8 & Ship & $[0,0,0,0,0,0,0,0,1,0]$ \\
9 & Truck & $[0,0,0,0,0,0,0,0,0,1]$ \\
\bottomrule
\end{tabular}
\end{table}

The separation of training, validation, and test data is essential for avoiding overly optimistic performance estimates. Training data determine the weights. Validation data guide design choices such as epoch count, architecture depth, and regularization strength. Test data provide the final estimate of how the selected model behaves on unseen inputs. Without the separation, a classifier could appear successful because it memorized the examples used for evaluation rather than learned a transferable visual representation.

\section{Neural Network Formulation}
\vspace{6pt}
\subsection{Fully Connected Multilayer Perceptron}
\vspace{6pt}
A multilayer perceptron consists of an input layer, one or more hidden layers, and an output layer. In a dense layer, every unit in the current layer receives a weighted combination of all activations from the previous layer. For layer $\ell$, the forward propagation equations are
\begin{align}
z^{(\ell)} &= W^{(\ell)}a^{(\ell-1)} + b^{(\ell)}, \\
a^{(\ell)} &= \phi\!\left(z^{(\ell)}\right),
\label{eq:mlp_forward}
\end{align}
where $W^{(\ell)}$ is the weight matrix, $b^{(\ell)}$ is the bias vector, $a^{(\ell-1)}$ is the previous activation, and $\phi(\cdot)$ is a nonlinear activation. Rectified linear units (ReLU) are widely used because they reduce saturation and support efficient gradient propagation \cite{nair2010rectified}. Weight initialization also affects stability, and variance-preserving initializers reduce the risk of exploding or vanishing activations in deep networks \cite{glorot2010understanding,he2015delving}.

\begin{figure}[t]
\centering
\resizebox{0.98\columnwidth}{!}{%
\begin{tikzpicture}[>=Stealth, neuron/.style={circle,draw,minimum size=6mm,inner sep=0pt,fill=gray!10}, layerlabel/.style={align=center,font=\footnotesize}]
\foreach \y in {0,1,2,3,4} {\node[neuron] (I\y) at (0,-\y*0.7) {};}
\foreach \y in {0,1,2,3} {\node[neuron] (H1\y) at (2.0,-0.35-\y*0.7) {};}
\foreach \y in {0,1,2,3} {\node[neuron] (H2\y) at (4.0,-0.35-\y*0.7) {};}
\foreach \y in {0,1,2} {\node[neuron] (O\y) at (6.0,-0.7-\y*0.7) {};}
\foreach \i in {0,1,2,3,4} {\foreach \j in {0,1,2,3} {\draw[gray!65] (I\i) -- (H1\j);}}
\foreach \i in {0,1,2,3} {\foreach \j in {0,1,2,3} {\draw[gray!65] (H1\i) -- (H2\j);}}
\foreach \i in {0,1,2,3} {\foreach \j in {0,1,2} {\draw[gray!65] (H2\i) -- (O\j);}}
\node[layerlabel] at (0,0.55) {Input\\$3072$ pixels};
\node[layerlabel] at (2.0,0.55) {Hidden\\dense + ReLU};
\node[layerlabel] at (4.0,0.55) {Hidden\\dense + ReLU};
\node[layerlabel] at (6.0,0.55) {Output\\10 classes};
\draw[decorate,decoration={brace,amplitude=5pt},xshift=-6pt] (-0.45,0.25) -- (-0.45,-2.8) node[midway,left=7pt,font=\footnotesize] {Flattened RGB vector};
\node[font=\footnotesize,align=center] at (3,-3.6) {Dense connections learn global correlations but do not explicitly preserve image locality.};
\end{tikzpicture}}
\caption{Fully connected multilayer perceptron used as a conceptual baseline for CIFAR-10. Vectorization produces a 3072-dimensional input; dense layers learn nonlinear combinations before the ten-class output layer.}
\label{fig:mlp}
\end{figure}

The dense architecture in Fig.~\ref{fig:mlp} can approximate complicated decision boundaries, but the parameterization is not specialized for images. A connection from a pixel in the top-left corner is treated like a connection from a neighboring pixel or from a distant corner. Therefore, the MLP must learn spatial relationships from data alone. Such behavior increases parameter demand and reduces sample efficiency compared with convolutional processing.

\subsection{Capacity and Parameter Efficiency}
Network capacity is determined not only by the number of layers but also by the way parameters are shared. For a dense layer with input dimension $d$ and hidden width $h$, the parameter count is:
\begin{equation}
N_{\mathrm{dense}}=(d+1)h,
\label{eq:dense_params}
\end{equation}
where the additional term accounts for the bias parameters. With $d=3072$ and $h=512$, a single first hidden layer already contains more than 1.57 million parameters. The large count can be appropriate when sufficient data are available, but it also increases memory cost and overfitting risk.

For a convolutional layer with $F$ filters, $C$ input channels, and an $r\times r$ kernel, the parameter count is
\begin{equation}
N_{\mathrm{conv}}=(r^2C+1)F.
\label{eq:conv_params}
\end{equation}
A $3\times3$ convolution with $C=3$ and $F=64$ has only 1,792 parameters because the same kernel weights are reused at every spatial position. The reduction is not merely computational; it encodes the assumption that local visual patterns such as edges and corners can appear anywhere in an image. Parameter sharing therefore improves sample efficiency and provides a theoretical reason for the empirical success of CNNs on small image datasets \cite{goodfellow2016deep}.

The distinction also clarifies why MLP and CNN results should not be interpreted only through final accuracy. A dense classifier tests whether global nonlinear combinations can separate classes after flattening. A convolutional classifier tests whether learned local features and hierarchical composition improve generalization. For CIFAR-10, the second formulation is better aligned with the statistical structure of natural images.
\subsection{Softmax Output and Cross-Entropy Loss}
The final layer produces logits $s\in\mathbb{R}^{10}$. Class probabilities are obtained with the softmax function
\begin{equation}
\hat{p}_{k}=\frac{\exp(s_k)}{\sum_{j=1}^{10}\exp(s_j)}, \qquad k=1,\ldots,10.
\label{eq:softmax}
\end{equation}
For a mini-batch of $B$ samples, the categorical cross-entropy loss is
\begin{equation}
\mathcal{L}=-\frac{1}{B}\sum_{i=1}^{B}\sum_{k=1}^{10} y_{i,k}\log\left(\hat{p}_{i,k}+\epsilon\right),
\label{eq:crossentropy}
\end{equation}
where $\epsilon$ is a small constant used for numerical stability. Minimizing \eqref{eq:crossentropy} increases the predicted probability assigned to the correct class and decreases probabilities assigned to incorrect classes. Back-propagation computes gradients of the loss with respect to all trainable parameters, and an optimizer uses those gradients to update the weights.

\section{Training Protocol and Hyperparameters}
\subsection{Gradient-Based Learning}
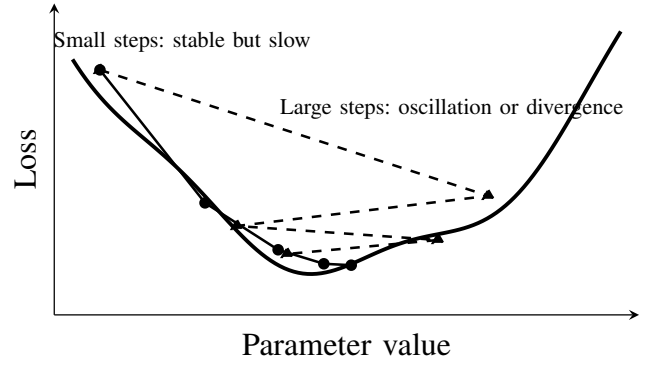
\begin{figure}[t]
\centering
\resizebox{0.95\columnwidth}{!}{%
\begin{tikzpicture}
\begin{axis}[
width=8.0cm,height=5.0cm,
axis lines=left,
xlabel={Parameter value},ylabel={Loss},
xtick=\empty,ytick=\empty,
ymin=-0.15,ymax=4.3,xmin=-3.2,xmax=3.2,
clip=false]
\addplot[domain=-3:3,samples=120,very thick] {0.35*x^2 + 0.20*sin(deg(2.6*x)) + 0.55};
\addplot[mark=*,mark size=1.4pt,thick] coordinates {(-2.7,3.35) (-1.55,1.45) (-0.75,0.78) (-0.25,0.58) (0.05,0.56)};
\addplot[mark=triangle*,mark size=1.8pt,thick,dashed] coordinates {(-2.7,3.35) (1.55,1.55) (-1.20,1.12) (1.00,0.92) (-0.65,0.72)};
\node[align=left,font=\scriptsize] at (axis cs:-1.8,3.75) {Small steps:\ stable\ but slow};
\node[align=left,font=\scriptsize] at (axis cs:1.15,2.8) {Large steps:\ oscillation\ or divergence};
\end{axis}
\end{tikzpicture}}
\caption{Conceptual effect of learning-rate selection during loss minimization. The learning rate controls the step size in parameter space and strongly affects convergence speed and stability.}
\label{fig:learning_rate}
\end{figure}

Let $\theta$ denote the collection of network weights and biases. In stochastic or mini-batch training, the parameter update has the generic form
\begin{equation}
\theta_{t+1}=\theta_t-\eta\,g_t,
\label{eq:gd}
\end{equation}
where $g_t$ is an estimate of the gradient at iteration $t$ and $\eta$ is the learning rate. A small learning rate produces stable but slow progress. A large learning rate can overshoot narrow minima or produce oscillatory behavior. Fig.~\ref{fig:learning_rate} visualizes the qualitative effect. Adaptive methods such as Adagrad and Adam rescale updates by using historical gradient statistics, improving practical training stability for many neural-network problems \cite{duchi2011adaptive,kingma2015adam}.

The experiment uses Adam with $\eta=0.001$, a value commonly adopted as a stable initial choice for deep learning experiments. Adam maintains first- and second-moment estimates of the gradients,
\begin{align}
m_t &= \beta_1m_{t-1}+(1-\beta_1)g_t,\\
v_t &= \beta_2v_{t-1}+(1-\beta_2)g_t^2,
\end{align}
and applies bias-corrected updates to each parameter. The method combines momentum-like smoothing with coordinate-wise scaling, allowing efficient progress even when gradient magnitudes differ across layers.

\subsection{Mini-Batch Size and Validation Monitoring}
The mini-batch size is set to 128. Mini-batches reduce the variance of gradient estimates compared with single-sample updates while retaining better hardware efficiency than full-batch optimization. Powers of two are often selected because they map efficiently to memory layouts and accelerator kernels. However, batch size is not merely an implementation detail. Very small batches can inject excessive gradient noise, whereas very large batches can reduce the beneficial stochasticity of training and may require learning-rate adjustment.

Validation monitoring is required because training loss alone cannot determine generalization. A decreasing training loss indicates improved fit to the training samples. A decreasing validation loss indicates improved expected performance on unseen samples drawn from the same distribution. When training loss decreases while validation loss increases, the model has begun to fit sample-specific patterns or noise. Early stopping, weight decay, dropout, data augmentation, and batch normalization are standard remedies \cite{srivastava2014dropout,ioffe2015batch,shorten2019survey}.

\subsection{Implementation and Reproducibility Considerations}
Reproducibility requires precise reporting of the data split, preprocessing operations, model configuration, optimizer, batch size, and number of epochs. Table~\ref{tab:config} summarizes the experimental configuration. The implementation can be reproduced in modern deep-learning environments such as PyTorch, which provides automatic differentiation, GPU support, and modular neural-network layers \cite{paszke2019pytorch}. Comparable results require the same normalization rule, the same validation protocol, and consistent logging of loss and accuracy after each epoch.

\begin{table}[t]
\caption{Reported Experimental Configuration}
\label{tab:config}
\centering
\footnotesize
\begin{tabular}{@{}ll@{}}
\toprule
Component & Configuration \\
\midrule
Dataset & CIFAR-10 RGB images, ten balanced classes \\
Input size & $32\times32\times3$ pixels \\
Preprocessing & Pixel normalization by division by 255 \\
Label format & Ten-dimensional one-hot vectors \\
Training split & CIFAR-10 training subset after validation holdout \\
Validation split & 5,000 images reserved from the training partition \\
Optimizer & Adam \\
Learning rate & 0.001 \\
Mini-batch size & 128 \\
Epochs & 10 \\
CNN structure & Six convolutional and three max-pooling layers \\
Output activation & Softmax over ten classes \\
Evaluation metrics & Cross-entropy loss and validation accuracy \\
\bottomrule
\end{tabular}
\end{table}

The use of a validation holdout is especially important because the experiment is short and the architecture has enough capacity to memorize patterns quickly. A reproducible protocol should store the random seed used for splitting, the initialization rule, and the mini-batch ordering. Reporting only the final validation accuracy is insufficient for scientific comparison; the full learning curve is needed to identify the epoch at which validation loss begins to increase.

A rigorous extension would repeat the same configuration over several random seeds and report the mean and standard deviation of validation accuracy. Such reporting would separate genuine architectural improvements from random variation caused by initialization or mini-batch order. The requirement is modest for CIFAR-10 because training time is relatively low, making repeated runs feasible in an academic laboratory setting.
\section{Convolutional Feature Learning}
\subsection{Convolutional Inductive Bias}
A convolutional layer computes feature maps by sliding trainable kernels over local neighborhoods. For an input tensor $X$ and a kernel $K$, a simplified single-output-channel convolution can be expressed as
\begin{equation}
Z(u,v)=\sum_{c=1}^{C}\sum_{i=1}^{r}\sum_{j=1}^{r} K(c,i,j)X(c,u+i,v+j)+b,
\label{eq:convolution}
\end{equation}
where $C$ is the number of input channels and $r\times r$ is the kernel size. The same kernel is reused across spatial locations, leading to weight sharing. The resulting representation is translation-sensitive at early layers and increasingly abstract at deeper layers. Such inductive bias explains why CNNs are usually more effective than MLPs for image classification when training data are limited \cite{lecun1998gradient,krizhevsky2012imagenet}.

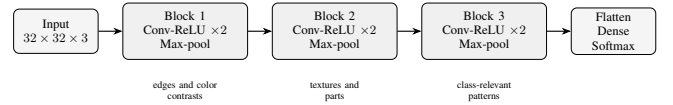
\begin{figure}[t]
\centering
\resizebox{0.96\columnwidth}{!}{%
\begin{tikzpicture}[>=Stealth,
block/.style={draw,rounded corners,minimum height=11mm,minimum width=21mm,align=center,font=\small,fill=gray!8},
group/.style={draw,rounded corners,minimum height=14mm,minimum width=31mm,align=center,font=\small,fill=gray!12},
arrow/.style={->,thick}]
\node[block] (in) {Input\\$32\times32\times3$};
\node[group,right=6mm of in] (g1) {Block 1\\Conv-ReLU $\times2$\\Max-pool};
\node[group,right=6mm of g1] (g2) {Block 2\\Conv-ReLU $\times2$\\Max-pool};
\node[group,right=6mm of g2] (g3) {Block 3\\Conv-ReLU $\times2$\\Max-pool};
\node[block,right=6mm of g3] (out) {Flatten\\Dense\\Softmax};
\foreach \a/\b in {in/g1,g1/g2,g2/g3,g3/out}{\draw[arrow] (\a) -- (\b);}
\node[font=\scriptsize,align=center,below=4mm of g1] {edges and color\\contrasts};
\node[font=\scriptsize,align=center,below=4mm of g2] {textures and\\parts};
\node[font=\scriptsize,align=center,below=4mm of g3] {class-relevant\\patterns};
\end{tikzpicture}}
\caption{Convolutional architecture evaluated in the experiment. The six convolutional layers are grouped into three Conv-ReLU pairs, and each group is followed by max-pooling before dense softmax classification.}
\label{fig:cnn_arch}
\end{figure}

The evaluated CNN contains six convolutional layers and three max-pooling layers, followed by flattening and dense classification. ReLU activations introduce nonlinearity after convolutional transformations. The output layer uses softmax to form a probability distribution over the ten categories. The organization is summarized in Fig.~\ref{fig:cnn_arch}. Early layers are expected to capture edges, color transitions, and small textures; intermediate layers combine those primitives into object parts; final layers produce class-discriminative features for the dense classifier.

\subsection{Pooling and Spatial Downsampling}
\begin{figure}[t]
\centering
\resizebox{0.86\columnwidth}{!}{%
\begin{tikzpicture}[cell/.style={draw,minimum width=7mm,minimum height=7mm,align=center,font=\scriptsize}, outcell/.style={draw,minimum width=7mm,minimum height=7mm,align=center,font=\scriptsize,fill=gray!12}, >=Stealth]
\matrix[matrix of nodes,ampersand replacement=\&,nodes=cell,row sep=-\pgflinewidth,column sep=-\pgflinewidth] (A) {
2 \& 2 \& 7 \& 3\\
9 \& 4 \& 6 \& 1\\
8 \& 5 \& 2 \& 4\\
3 \& 1 \& 2 \& 6\\};
\node[draw,rounded corners,fit=(A-1-1)(A-2-2),inner sep=0pt,very thick] {};
\node[draw,rounded corners,fit=(A-1-3)(A-2-4),inner sep=0pt,very thick] {};
\node[draw,rounded corners,fit=(A-3-1)(A-4-2),inner sep=0pt,very thick] {};
\node[draw,rounded corners,fit=(A-3-3)(A-4-4),inner sep=0pt,very thick] {};
\node[right=12mm of A] (label) {\footnotesize $2\times2$ max-pool, stride 2};
\draw[->,thick] (A.east) -- (label.west);
\matrix[matrix of nodes,ampersand replacement=\&,nodes=outcell,row sep=-\pgflinewidth,column sep=-\pgflinewidth,right=12mm of label] (B) {
9 \& 7\\
8 \& 6\\};
\draw[->,thick] (label.east) -- (B.west);
\end{tikzpicture}}
\caption{Max-pooling example. Each nonoverlapping $2\times2$ region is replaced by its largest activation, reducing the feature-map size while retaining strong local evidence.}
\label{fig:maxpool}
\end{figure}
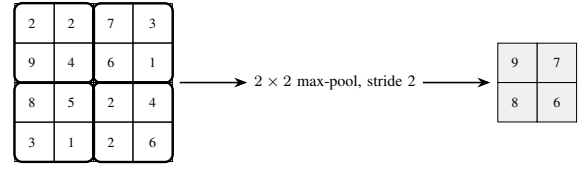

Pooling reduces the spatial dimensions of a feature map while preserving prominent activations. Max-pooling with a $2\times2$ window and stride 2 is defined as
\begin{equation}
P(u,v)=\max_{0\le i<2,\,0\le j<2} Z(2u+i,2v+j).
\label{eq:maxpool}
\end{equation}
Fig.~\ref{fig:maxpool} shows a numerical example. Pooling increases the receptive field of later layers and reduces memory and computation. It also provides limited tolerance to small translations because the maximum activation remains stable when the strongest local feature shifts slightly within the pooling window.

Pooling also has limitations. Aggressive downsampling can discard location information that may be useful for fine-grained classes. For CIFAR-10, the input image is already small, and therefore the number and placement of pooling stages must be selected carefully. Three pooling stages reduce spatial resolution substantially; adding more stages would risk eliminating useful structure before sufficient high-level features are computed.

\section{Experimental Results}
\subsection{Training Log}
The CNN is trained for ten epochs. Table~\ref{tab:training_log} reports the loss and validation accuracy recorded at each epoch. The values indicate rapid improvement during the first four epochs, followed by slower validation gains. The best validation accuracy is observed near epoch 6, with only small changes thereafter. Meanwhile, the validation loss decreases until approximately epoch 4 or 6 and then increases, while the training loss decreases monotonically. The divergence between the two losses is a typical signal that additional epochs increase specialization to the training set.

\begin{table}[t]
\caption{Training and Validation Log for the CNN Classifier}
\label{tab:training_log}
\centering
\footnotesize
\begin{tabular}{@{}cccc@{}}
\toprule
Epoch & Training loss & Validation loss & Validation accuracy \\
\midrule
0 & 1.8060 & 1.4998 & 0.4401 \\
1 & 1.2761 & 1.1801 & 0.5688 \\
2 & 0.9896 & 0.9389 & 0.6694 \\
3 & 0.8062 & 0.8489 & 0.7079 \\
4 & 0.6631 & 0.7718 & 0.7367 \\
5 & 0.5319 & 0.8143 & 0.7340 \\
6 & 0.4052 & 0.7733 & 0.7477 \\
7 & 0.2986 & 0.8657 & 0.7445 \\
8 & 0.2253 & 0.9494 & 0.7431 \\
9 & 0.1598 & 1.1255 & 0.7360 \\
\bottomrule
\end{tabular}
\end{table}

The improvement from 44.01\% validation accuracy at epoch 0 to 74.77\% at epoch 6 demonstrates that the model learns meaningful visual features within a small number of passes through the data. The decrease at epoch 9 indicates that higher training accuracy is not necessarily equivalent to better generalization. In low-resolution object recognition, misleading details and background correlations can be learned by a sufficiently flexible model. Validation monitoring therefore becomes a necessary part of experimental interpretation.

\begin{figure*}[t]
\centering
\begin{minipage}{0.48\textwidth}
\centering
\begin{tikzpicture}
\begin{axis}[
width=\linewidth,height=5.0cm,
xlabel={Epoch},ylabel={Validation accuracy},
ymin=0.40,ymax=0.78,xmin=0,xmax=9,
grid=both,minor tick num=1,legend pos=south east]
\addplot[mark=*,thick] coordinates {(0,0.4401) (1,0.5688) (2,0.6694) (3,0.7079) (4,0.7367) (5,0.7340) (6,0.7477) (7,0.7445) (8,0.7431) (9,0.7360)};
\legend{Validation}
\end{axis}
\end{tikzpicture}
\end{minipage}\hfill
\begin{minipage}{0.48\textwidth}
\centering
\begin{tikzpicture}
\begin{axis}[
width=\linewidth,height=5.0cm,
xlabel={Epoch},ylabel={Loss},
ymin=0.0,ymax=1.95,xmin=0,xmax=9,
grid=both,minor tick num=1,legend pos=north east]
\addplot[mark=*,thick] coordinates {(0,1.8060) (1,1.2761) (2,0.9896) (3,0.8062) (4,0.6631) (5,0.5319) (6,0.4052) (7,0.2986) (8,0.2253) (9,0.1598)};
\addplot[mark=square*,thick,dashed] coordinates {(0,1.4998) (1,1.1801) (2,0.9389) (3,0.8489) (4,0.7718) (5,0.8143) (6,0.7733) (7,0.8657) (8,0.9494) (9,1.1255)};
\legend{Training,Validation}
\end{axis}
\end{tikzpicture}
\end{minipage}
\caption{Validation accuracy and loss curves redrawn from the experimental log. Accuracy stabilizes near 75\%, while validation loss increases after the middle epochs even though training loss continues to decrease.}
\label{fig:curves}
\end{figure*}

\subsection{Interpretation of Learning Curves}
Fig.~\ref{fig:curves} presents a clearer visualization of the training dynamics. The accuracy curve rises sharply in the early epochs and then enters a plateau. The loss plot provides additional information: training loss continues to fall from 1.8060 to 0.1598, but validation loss reaches a low range around epochs 4--6 and then rises to 1.1255. The pattern indicates that the model becomes increasingly confident on the training set while its probabilistic predictions on validation data become less calibrated. A model can therefore maintain similar validation accuracy while assigning excessive confidence to wrong validation predictions, producing a larger validation loss.

The result suggests that epoch selection should not be based solely on accuracy. Cross-entropy loss is sensitive to confidence and penalizes confident incorrect decisions strongly. Therefore, a validation-loss minimum often provides a better stopping point than a late epoch with marginally similar accuracy. In the recorded experiment, continuing from epoch 6 to epoch 9 reduces training loss by more than 60\%, yet validation accuracy decreases from 74.77\% to 73.60\%. The gap supports early stopping around the best validation region.

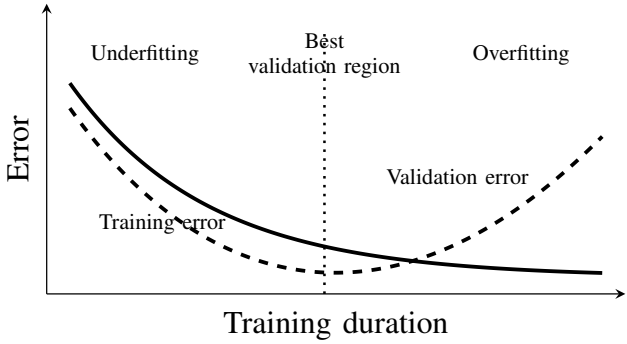
\begin{figure}[t]
\centering
\resizebox{0.94\columnwidth}{!}{%
\begin{tikzpicture}
\begin{axis}[
width=8.0cm,height=4.8cm,
axis lines=left,
xlabel={Training duration},ylabel={Error},
xtick=\empty,ytick=\empty,
ymin=0,ymax=3.1,xmin=0,xmax=10,clip=false]
\addplot[domain=0.4:9.6,samples=120,very thick] {2.45*exp(-0.42*x)+0.18};
\addplot[domain=0.4:9.6,samples=120,very thick,dashed] {0.18+0.065*(x-4.8)^2+0.68*exp(-0.55*x)};
\draw[dotted,thick] (axis cs:4.8,0) -- (axis cs:4.8,2.8);
\node[font=\scriptsize,align=center] at (axis cs:1.7,2.55) {Underfitting};
\node[font=\scriptsize,align=center] at (axis cs:4.8,2.55) {Best\\validation region};
\node[font=\scriptsize,align=center] at (axis cs:8.2,2.55) {Overfitting};
\node[font=\scriptsize] at (axis cs:2.0,0.75) {Training error};
\node[font=\scriptsize] at (axis cs:7.1,1.25) {Validation error};
\end{axis}
\end{tikzpicture}}
\caption{Generalization behavior reflected by training and validation errors. The optimal stopping region occurs before validation error rises due to overfitting.}
\label{fig:generalization}
\end{figure}

The conceptual relation in Fig.~\ref{fig:generalization} explains why additional capacity or epochs must be balanced by regularization. Underfitting occurs when the model is too simple or has not trained long enough to capture the main patterns. Overfitting occurs when the model captures training-specific fluctuations that do not transfer to validation or test data. The observed curves place the evaluated CNN near the transition: a useful representation has been learned, but generalization starts to degrade as optimization continues.

\subsection{MLP and CNN Comparison}
A dense MLP remains a useful reference architecture because it demonstrates the baseline capability of nonlinear supervised learning. Nevertheless, CIFAR-10 classification requires spatial feature extraction. A dense network processes a flattened input vector and must learn edge detectors, local textures, and spatial hierarchies implicitly through its weights. A CNN encodes locality and translation-related structure directly through its architecture. Consequently, convolutional models usually reach stronger image-classification performance with fewer effective parameters and better sample efficiency \cite{lecun1998gradient,krizhevsky2012imagenet,he2016deep}.

The experimental behavior supports the suitability of convolutional architectures for image classification. The CNN reaches a validation accuracy close to 75\% after ten epochs without advanced regularization or pretrained features. The performance is reasonable for a compact architecture and a small input resolution. The remaining error indicates that object ambiguity, limited resolution, and class similarity still impose substantial difficulty. Cats and dogs, trucks and automobiles, and birds and airplanes are common sources of confusion in CIFAR-like settings because the discriminative information can occupy only a few pixels.

\section{Discussion and Future Research Importance}
\subsection{Experimental Value}
The experiment has value beyond the single reported accuracy value. First, it demonstrates the entire supervised-learning pipeline from preprocessing to validation analysis. Second, it shows why loss curves and accuracy curves must be examined together. Third, it exposes the practical consequences of architecture choice: dense layers can perform nonlinear classification, but convolutional feature extraction is more appropriate when image locality matters. Fourth, it produces a reproducible baseline that future studies can improve in controlled increments.

For educational research, the setup is suitable for laboratory assignments in which students compare MLPs, shallow CNNs, and deeper CNNs under equal data splits. For engineering research, the same baseline can be extended with regularization and augmentation to quantify the contribution of each design decision. For applied research, the low-resolution setting resembles edge-device or bandwidth-limited visual sensing, where compact models remain important.

\subsection{Recommended Extensions}
Several extensions can improve the reported baseline. Data augmentation can create transformed training images through random cropping, horizontal flipping, color jittering, and cutout-like occlusion. Such transformations encourage invariance and reduce dependence on accidental backgrounds \cite{shorten2019survey}. Mixup and automated augmentation-policy search offer stronger distribution-expansion strategies for future CIFAR-10 trials \cite{zhang2018mixup,cubuk2019autoaugment}. Dropout can reduce co-adaptation among dense units by randomly disabling activations during training \cite{srivastava2014dropout}. Batch normalization can stabilize intermediate activation distributions and permit faster training with less sensitivity to initialization \cite{ioffe2015batch}. Weight decay can penalize overly large parameters and improve smoothness of the learned function; decoupled weight decay is a relevant extension when Adam is retained as the optimizer \cite{loshchilov2019decoupled}.

Architecture design can also be improved. Modern CNNs such as AlexNet and residual networks demonstrated that depth and skip connections can improve representation learning at scale \cite{krizhevsky2012imagenet,he2016deep}. Very deep small-filter networks and densely connected models provide additional comparison points for future controlled experiments \cite{simonyan2015very,huang2017densely}. For CIFAR-10, a residual architecture would allow deeper feature extraction while reducing optimization difficulty. Transfer learning is another future direction, although small $32\times32$ images require careful resizing and preprocessing when pretrained models are used. Hyperparameter search over learning rate, batch size, optimizer, number of channels, kernel size, and regularization coefficients would produce a stronger empirical conclusion than a single configuration.

Evaluation should also move beyond global accuracy. A confusion matrix would reveal which classes are most frequently confused. Per-class precision, recall, and F1-score would distinguish balanced performance from category-specific weakness. Calibration metrics would determine whether predicted probabilities reflect actual correctness frequencies. Finally, repeated runs with different random seeds would quantify statistical variability, which is important because neural-network training is influenced by initialization, data ordering, and mini-batch composition.

\subsection{Implications for Future Studies}
Future studies can use the presented baseline to evaluate how architectural inductive bias and regularization interact. A promising sequence is to begin with the six-convolutional-layer model, add data augmentation, introduce dropout or weight decay, add batch normalization, and then compare a residual architecture under the same split. Such a sequence would make the contribution of each improvement measurable. The recorded overfitting behavior also makes the experiment suitable for studying early stopping and validation-based model selection.

The experiment is important for future low-resource visual-recognition systems because CIFAR-10 forces models to make decisions under severe spatial compression. Many real deployments, including embedded cameras and communication-constrained sensors, encounter similarly limited resolution or restricted computation. A compact CNN that reaches moderate accuracy with few epochs provides a practical starting point for further compression, pruning, quantization, and deployment studies. Therefore, the experimental baseline can support future research connecting machine-learning theory, educational reproducibility, and efficient computer-vision implementation.

\subsection{Future Experimental Protocol}
A stronger follow-up protocol should be organized as a sequence of ablation experiments rather than as an isolated accuracy measurement. The baseline should first be trained under a fixed split and fixed seed. Each later experiment should introduce only one additional component, such as augmentation, mixup, dropout, decoupled weight decay, batch normalization, or a residual block. The resulting comparison would identify whether an accuracy gain is produced by additional capacity, improved regularization, or better optimization conditioning \cite{srivastava2014dropout,ioffe2015batch,he2016deep,shorten2019survey,zhang2018mixup,loshchilov2019decoupled}.

Future evaluation should include accuracy, loss, class-wise precision, class-wise recall, F1-score, a normalized confusion matrix, and repeated runs over several random seeds. These metrics would reveal whether errors are concentrated in visually related categories such as cat and dog or automobile and truck, and they would separate robust methodological improvement from incidental variation caused by initialization, data ordering, or hardware-dependent implementation details.

\subsection{Limitations and Threats to Validity}
Several limitations should be considered when interpreting the results. The reported training log corresponds to one configuration and one validation split; therefore, the numerical accuracy should be treated as a baseline rather than a definitive performance limit. Exact replication would also require the full filter configuration, initialization seed, and mini-batch order. The absence of a confusion matrix restricts class-level interpretation, while the absence of strong regularization and augmentation means that the reported accuracy should not be interpreted as the highest achievable CIFAR-10 performance. The value of the baseline lies in clarity: the learning dynamics remain easy to inspect, and each future modification can be measured against a transparent starting point.
\section{Conclusion}
A formal reconstruction of CIFAR-10 image classification using neural networks has been developed with an emphasis on theoretical clarity, visual explanation, and experimental interpretation. The MLP formulation illustrates dense nonlinear classification after pixel vectorization, whereas the CNN formulation demonstrates the value of local receptive fields, shared filters, and pooling for visual data. The evaluated convolutional architecture reaches approximately 74.77\% validation accuracy under a ten-epoch training protocol with Adam optimization and a learning rate of 0.001.

The training log shows that validation accuracy plateaus while validation loss begins to rise after the middle epochs. The behavior indicates that continued optimization improves training-set fit but reduces probabilistic generalization on validation data. Early stopping, regularization, augmentation, and deeper architectures are therefore appropriate next steps. The experimental baseline provides a useful academic reference for future studies on CIFAR-10, low-resolution image recognition, and reproducible neural-network education.

The documented results also establish a foundation for controlled comparison with future compact vision models. Because the data split, optimizer, batch size, epoch count, and validation behavior are explicitly reported, later experiments can evaluate improvements through reproducible ablation rather than through isolated final scores. Future investigations should preserve the same reporting discipline while adding augmentation, regularization, and alternative CNN architectures one component at a time. Such a protocol would clarify whether improved accuracy is produced by stronger feature extraction, better optimization, or reduced overfitting.

The significance of the experiment is therefore methodological as well as numerical. The learning curves, validation gap, and model-capacity analysis provide a transparent baseline for future studies on reliable model selection, efficient low-resolution recognition, class-level error diagnosis, and reproducible neural image-classification laboratories. The extended reference basis further connects the baseline to modern augmentation, optimization, and convolutional architecture research, enabling later studies to formulate stronger hypotheses before computationally expensive experimentation begins.

\enlargethispage{2\baselineskip}
\balance

\end{document}